\documentclass[10pt,twocolumn,letterpaper]{article}

\usepackage[pagenumbers]{cvpr} 

\usepackage{amsmath,amsfonts,bm}

\def\eqref#1{equation~\ref{#1}}

\def\1{\bm{1}}

\def\vf{{\bm{f}}}

\def\vn{{\bm{n}}}

\DeclareMathAlphabet{\mathsfit}{\encodingdefault}{\sfdefault}{m}{sl}
\SetMathAlphabet{\mathsfit}{bold}{\encodingdefault}{\sfdefault}{bx}{n}

\def\gD{{\mathcal{D}}}

\def\gF{{\mathcal{F}}}

\def\sR{{\mathbb{R}}}

\newcommand{\vx}{\boldsymbol{x}}
\newcommand{\vy}{\boldsymbol{y}}

\usepackage{enumitem}
\usepackage{xspace}
\usepackage{booktabs}
\usepackage{colortbl}
\usepackage{pifont}
\usepackage{graphicx}
\usepackage{siunitx}
\usepackage{multirow}

\usepackage{caption}
\usepackage{algorithm}
\usepackage{algpseudocode}
\usepackage{algorithmicx}
\algrenewcommand\alglinenumber[1]{\tiny\color{gray}{#1}}
\algrenewcommand\algorithmicindent{1.0em}

\usepackage{url}

\def\modelname{FOFPred\xspace}

\definecolor{Gray}{gray}{0.90}

\newcommand{\cmark}{\ding{51}}
\newcommand{\xmark}{\ding{55}}

\newcommand{\inc}[1]{\ensuremath{_{\text{\textcolor{PineGreen}{(+#1)}}}}}
\newcommand{\dec}[1]{\ensuremath{_{\text{\textcolor{RedOrange}{(-#1)}}}}}

\newcommand{\bhdr}[1]{\noindent\textbf{#1}}

\usepackage{tocloft}
\usepackage{etoc}
\addtocontents{toc}{\protect\setcounter{tocdepth}{-1}}

\definecolor{cvprblue}{rgb}{0.21,0.49,0.74}
\usepackage[pagebackref,breaklinks,colorlinks,allcolors=cvprblue]{hyperref}

\def\paperID{42574} 
\def\confName{CVPR}
\def\confYear{2026}

\title{
Future Optical Flow Prediction Improves Robot Control \& Video Generation
}

\author{%
Kanchana Ranasinghe$^{\star,1,2}$, 
Honglu Zhou$^1$, 
Yu Fang$^{\star,1}$, 
Luyu Yang$^1$, 
Le Xue$^1$, \vspace{0.1em} \\
Ran Xu$^1$, 
Caiming Xiong$^1$,
Silvio Savarese$^1$,
Michael S Ryoo$^{1,2}$,
Juan Carlos Niebles$^1$
\vspace{0.5em} \\
$^1$Salesforce AI Research \quad $^2$Stony Brook University \vspace{0.5em} \\
\texttt{\href{https://fofpred.github.io}{FOFPred.github.io}} \\
}

\begin{document}

\twocolumn[{%
\renewcommand\twocolumn[1][]{#1}%
\maketitle

\begin{center}
    \setcounter{figure}{0}
    \centering
    \captionsetup{type=figure}
    \vspace{-2.0em}
    \includegraphics[width=0.95\linewidth]{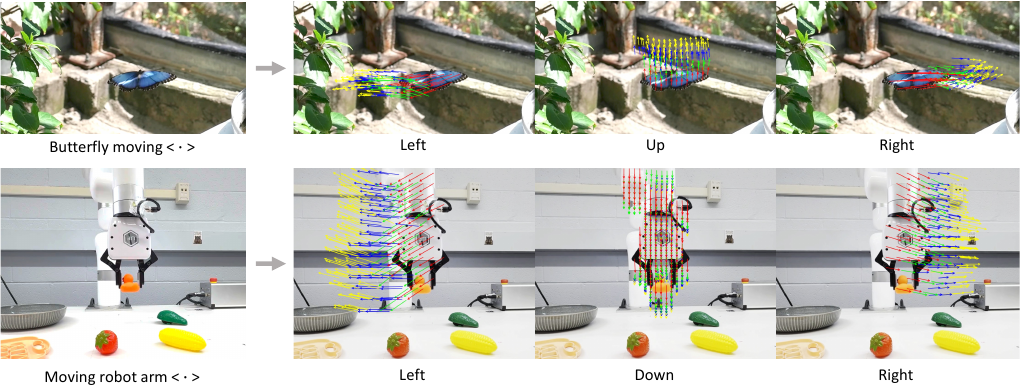}
    \vspace{-0.5em}
    \caption{
    \textbf{Our proposed model, \modelname, learns generalizable \textit{language-driven future optical flow prediction} that supports robot control and video generation.}
    $<\cdot >$ is a placeholder for text (e.g., Left/Up/Right). 
    For visualization purposes, we sample points from the predicted future optical flow 
    and show their predicted trajectories in this figure. 
    Checkout our \href{https://fofpred.github.io}{website} for more visualizations.
    }%
    \label{fig:teaser}%
    \vspace{0.8em}
\end{center}%
}]

\maketitle
\insert\footins{\noindent\footnotesize $^\star$Work done during an internship at Salesforce.}

\begin{abstract}
Future motion representations, such as optical flow, offer immense value for control and generative tasks. However, forecasting generalizable spatially dense motion representations remains a key challenge, and learning such forecasting from noisy, real-world data remains relatively unexplored. 
We introduce \modelname, a novel language-conditioned optical flow forecasting model featuring a unified Vision-Language Model (VLM) and Diffusion architecture. 
This unique combination enables strong multimodal reasoning with pixel-level generative fidelity for future motion prediction. 
Our model is trained on web-scale human activity data—a highly scalable but unstructured source. 
To extract meaningful signals from this noisy video-caption data, we employ crucial data preprocessing techniques and our unified architecture with strong image pretraining. 
The resulting trained model is then extended to tackle two distinct downstream tasks in control and generation. 
Evaluations across robotic manipulation and video generation under language-driven settings establish the cross-domain versatility of \modelname, confirming the value of a unified VLM-Diffusion architecture and scalable learning from diverse web data for future optical flow prediction.
\vspace{-0.5em}
\end{abstract}

\section{Introduction}

Recent studies have investigated the use of motion representations such as optical flow~\cite{Xu2024FlowAT,Ran25LangToMo}, sparse motion trajectories~\cite{yuan2024general,Wen2023AnypointTM,Bharadhwaj2024Track2ActPP}, and direction commands~\cite{Burgert2025GowiththeFlowMV,Wang2024MotionCtrl,yin2023dragnuwa} for multiple control and generation tasks. 
The main idea behind such motion representations, often in the form of \emph{future} pixel movements like optical flows and trajectories, is to explicitly capture desired dynamics necessary for downstream tasks such as robot control and video generation.

In the case of robot control, models with such future motion representations have the capability to explicitly consider future pixel displacement information to infer robot actions \cite{yuan2024general,Wen2023AnypointTM,Xu2024FlowAT,Bharadhwaj2024Track2ActPP,Ran25LangToMo}.
This is in contrast to the conventional Vision-Language-Action (VLA) models taking only RGB-frames and language instructions as input. 
Similarly, video diffusion models utilizing motion representations enables generation of videos with consistent and detailed movements, and it does it better than the counterparts utilizing language-only instructions \cite{Chefer2025VideoJAMJA,Liang2024MoVideoMV, Koroglu2024OnlyFlowOF,Gao2024FLIPFG}.

The key is in reliable and generalizable computation of such motion representation, which often involves \emph{future forecasting}. What has to be provided to the models for many (robot) control and (video) generation tasks is how things should move in the future, not what motion has been observed in the past. This inherently is a challenging task: it requires a clear formulation of what type of `future motion' we will be computing, curating large enough training data with strong supervisory signals, and an elegant model capable of fully digesting such (potentially cross-domain) training data while generalizing across domains.

In this paper, we formulate the problem as the forecasting of future \emph{optical flows} conditioned on language instructions, and present a new model architecture to forecast them. We also discuss how we can train our model with videos from various sources, ranging from human videos to robot videos. Optical flows are spatially dense motion representations, capturing movements of every pixel in the scene. This better preserves motion details compared to sparse representation, and enables easy use of diffusion models to learn their distributions in the form of images. In contrast to predicting RGB frame sequences, which also capture future motion \cite{Hu2024VideoPP}, optical flow eliminates static information irrelevant to motion, leading to more motion-centric representations \cite{Ran25LangToMo}.   

We introduce an optical flow forecasting model unifying both Vision-Language Models (VLM) and Diffusion. Fully taking advantage of VLM reasoning capability and pre-trained diffusion model image generation ability, we train our VLM-Diffusion model to generate future optical flow images. 
Our model is trained on web-scale human activity videos with paired captions allowing highly-scalable training. The resulting model, named \modelname, enables highly generalizable language-driven future optical flow prediction as illustrated in \Cref{fig:teaser}. 

While learning from web videos provides this strong generality, such data is highly unstructured with noisy videos and captions. 
Most prior work learning future motion representations either avoid training with such data \cite{yuan2024general,Wen2023AnypointTM,Xu2024FlowAT,Bharadhwaj2024Track2ActPP,Ran25LangToMo} or utilize RGB frame prediction \cite{Hu2024VideoPP,Ko2023LearningTA}. In \cite{Yang2025MagmaAF} where such training data is used, explicit camera motion correction is used to handle noisy videos and a VLM backbone enables learning from diverse noisy captions. However, the use of only a VLM restricts their future motion prediction to sparse pixel trajectories. 
In our work, we calculate dense relative optical flow compensating for the camera motion and model this with diffusion, which improves our learning from noisy web videos. 

Finally, we validate the effectiveness and generality of \modelname through evaluation on two challenging downstream tasks: language-driven robotic manipulation and language-guided motion video generation. For each task, we attach a diffusion policy head or a video diffusion head on top of \modelname and fine-tune the two heads separately for the respective downstream domains.

We summarize our key contributions as follows:
\begin{itemize}[leftmargin=1.5em,noitemsep,topsep=0.5ex,itemsep=0.0ex,partopsep=0ex,parsep=1ex]
    \item \textbf{Unified VLM-Diffusion:} Adopting this recent architecture for generalizable future optical flow prediction.
    
    \item \textbf{Scalable Learning:} Establish framework for learning future optical flow prediction from web-scale human activity videos.

    \item \textbf{Cross-Domain Versatility:} To the best of our knowledge, the first to utilize VLM–diffusion backbone for robotic manipulation and controlled video generation under language-driven settings.
\end{itemize}
\vspace{0.5em}

\begin{figure*}[t]
    \centering
    \includegraphics[width=0.90\linewidth]{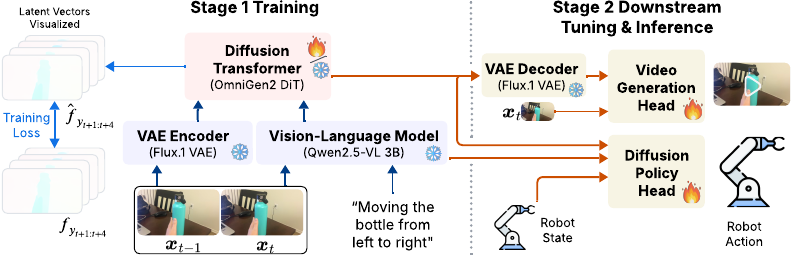}
    \caption{\textbf{Overview of Proposed \modelname:}
    (Left \& Center) We present the unified VLM-Diffusion architecture used in \modelname. Only the DiT module is trained while the VAE and VLM remain frozen. 
    (Right) We illustrate two distinct pipelines constructed with \modelname for two orthogonal tasks in control and generation. Each task specific head is first finetuned prior to inference on the downstream task. 
    } 
    \label{fig:arch}
\end{figure*}

\section{Related Work}

\bhdr{Unified Model for Control Generation.} 
Recent progress in foundation models has demonstrated a powerful trend toward unified model architectures that integrate diverse modalities and achieve highly controlled visual input manipulation. These generalist models \cite{Nvidia2025GR00TNA, Zhou2023UniDiffuserUT, Chen2025BLIP3oAF, Yang2025MagmaAF, Xiao2024OmniGenUI, Wu2025OmniGen2ET, Pan2025TransferBM, ju2025fulldit, gupta2025unified}, often building on transformers and large-scale pre-training (e.g., Gato \cite{DeepMind2022AGAF}), establish single frameworks capable of handling hundreds of tasks. In the generative domain, this unification enables advanced control: models like UniDiffuser \cite{Zhou2023UniDiffuserUT} tackle various image generation and editing tasks within a single conditional framework, while BLIP-3o \cite{Chen2025BLIP3oAF}, OmniGen-2 \cite{Wu2025OmniGen2ET}, and MetaQuery \cite{Pan2025TransferBM} demonstrate sophisticated language-conditioned iterative modification and transfer across modalities. Crucially, the development of plug-and-play architectures like ControlNet \cite{Zhang2023AddingCT} established the blueprint for adding flexible, low-level spatial controls to pretrained diffusion models. This pioneering work inspired recent unified generative backbones that accept diverse conditioning signals, enabling versatile controllable generation \cite{krishnan2025orchid, ju2025fulldit, gupta2025unified, liu2025video, hunyuan3d2025hunyuan3d, kant2025pippo, sun2025minimal, cho2025tc}. This foundation of multi-modal input, unified architecture, and flexible control is a key enabler for tackling complex tasks like controlled video \cite{peng2024controlnext, wang2025cinemaster, peng2024conditionvideo, tu2024motioneditor} and robotic action \cite{wang2025language, liu2024diff, li2025object, pinyoanuntapong2024controlmm} generation.

\bhdr{Motion Representations in Robot Control.}
Learning motion related information from videos has been widely explored in robot learning \citep{lee2017learning,finn2017deep,sun2018neural,kurutach2018learning,pari2021surprising,nair2022r3m,shao2021concept2robot,chen2021learning,bahl2022human,sharma2019third,du2023learning,sivakumar2022robotic,Sudhakar2024ControllingTW,Ko2023LearningTA,Hu2024VideoPP,Ren2025MotionTA}. Building on this, the field has increasingly explored vision-language-action (VLA) models to learn policies conditioned on natural language \citep{du2023learning,Sudhakar2024ControllingTW,Ko2023LearningTA,Hu2024VideoPP}, although many of these still rely on extensive action-trajectory annotations or task-specific heuristics. Recent foundation models aim to overcome this by better integrating spatio-temporal intelligence: MAGMA \cite{Yang2025MagmaAF} utilizes ``Trace-of-Mark'' annotations on video movements for action grounding and planning across digital and physical tasks. Other approaches focus on improving world model reasoning before action generation: FlowVLA \cite{Zhong2025FlowVLATI} employs a ``Visual Chain of Thought'' to explicitly predict optical flow before the next visual frame, disentangling motion dynamics from static appearance for more efficient policy learning, while DreamVLA \cite{Zhang2025DreamVLAAV} forecasts a compact set of crucial ``world knowledge'', including dynamic regions, depth, and semantics, to guide action planning and establish a robust perception-prediction-action (PPA) loop. Independently, optical flow has served as a primary representation in computer vision for motion learning \citep{Han2020SelfsupervisedCF,Sharma2022PixellevelCF,Luo2024FlowDiffuserAO,wang2024sea,gehrig2024dense}, and its extension to trajectories of pixel subsets \citep{yuan2024general,Wen2023AnypointTM,Xu2024FlowAT,Bharadhwaj2024Track2ActPP} has found utility in robot control. However, methods relying on localized trajectories \cite{yu2025objectmover,kwon2024language,cha2024text2hoi,song2025vl} often limit their focus to specific object movements, overlooking crucial global information, such as the overall movement of a manipulator. 
In contrast, our \modelname predicts spatially dense future optical flow leveraging powerful unified model architectures. 

\bhdr{Motion Control in Video Generation.}
Controlling motion in video generation using spatial conditioning signals has been widely explored \cite{Burgert2025GowiththeFlowMV,Zhao2024DartControlAD,Geng2024MotionPC,Shi2024MotionI2VCA,Wang2024MotionCtrl,yin2023dragnuwa,chen2023motionconditioned,Li2024ImageConductor,Wu2024DragAnything,Niu2024MOFAVideo,Wang2023VideoComposer,Zhang2024Tora,Zhou2024TrackGo,Lei2025AnimateAnything,hao2018controllable}. Some methods define object motion through sparse user-provided trajectories and points, including DragNUWA \cite{yin2023dragnuwa}, DragAnything \cite{Wu2024DragAnything}, Tora \cite{Zhang2024Tora}, TrackGo \cite{Zhou2024TrackGo}, and the foundational Controllable Video Generation \cite{hao2018controllable}. Others employ explicit motion models, such as MotionCtrl \cite{Wang2024MotionCtrl}, Motion-I2V \cite{Shi2024MotionI2VCA}, and Motion-Conditioned Diffusion Models \cite{chen2023motionconditioned}. Additional techniques achieve control through motion fields (MOFA-Video \cite{Niu2024MOFAVideo}), image-based precision guidance (Image Conductor \cite{Li2024ImageConductor}), compositional video synthesis (VideoComposer \cite{Wang2023VideoComposer}), or generalized animation (AnimateAnything \cite{Lei2025AnimateAnything}). However, language-based explicit motion control remains underexplored, which we focus on in this work.

\section{Method}

Optical Flow, the apparent displacement of pixels between frame pairs, has a long history in computer vision \cite{robot_vision,Barron1992PerformanceOO}. 
Estimating the true optical flow given a frame pair is a well-established problem \cite{teed2020raft}.
On the other hand, several recent works explore a different task of estimating future optical flow from a \textit{single frame} \cite{Walker2015DenseOF,Gao2017Im2FlowMH,Aleotti2021LearningOF}, with some conditioned on language \cite{Xu2024FlowAT,Ran25LangToMo}. To distinguish this task of future pixel displacement estimation from a single frame, we use the term future optical flow in the following sections. 

First, we introduce \modelname, our framework that generates future optical flow, given a language instruction and one or more images.
This future optical flow generation is learned from video-caption pairs without any pixel-level human annotation. 
We train a unified LLM-diffusion architecture based model to generate sequences of future optical flow frames. 
In the following sections, we first detail the architecture of our framework, followed by training and inference procedures. 
Next we describe our extension for robot control, a task requiring strong awareness of object and ego motions. 
Finally we present our second extension for motion control in text-to-video generation.

\subsection{Architecture}
\label{subsec:arch}
Motivated by recent success of unified model architectures in language conditioned visual generation \cite{Chen2025BLIP3oAF,Xiao2024OmniGenUI,Wu2025OmniGen2ET,Pan2025TransferBM}, we construct \modelname that maps frame sequences and language into future optical flow sequences. We illustrate an architectural overview of \modelname in \Cref{fig:arch}. 

Consider scene observations (images) $\vx_t \in \sR^{h \times w \times c}$ corresponding to time step $t$ and paired natural language caption $c$. 
We first utilize an autoregressive transformer based vision-language model as a textual feature encoder that encodes both the language caption $c$ and visual input $\vx_{t-1}, \vx_t$ to obtain $\vf_c$. 
We next encode the visual inputs, $\vx_{t-1}, \vx_t$ with a VAE encoder to obtain $\vf_v$.   
The textual ($\vf_c$) and visual ($\vf_v$) features are passed through MLP layers and input to our diffusion transformer (DiT). 
The outputs of the diffusion transformer, $\hat{\vf}_y$ are passed through a VAE decoder to obtain $\hat{\vy}$ sequences which correspond to future optical flow sequences.

For our DiT architecture, we utilize the OmniGen transformer \cite{Wu2025OmniGen2ET}, introducing several modifications to enable temporal modeling, given the time dimension of our framework's inputs and outputs. 
We first modify the DiT 2D RoPE encoding to handle both input and output frame sequences.
Next we update the DiT transformer blocks to perform full spatio-temporal attention in order to model the temporal axes in frame sequences. 
This design introduces no additional learnable parameters, enabling our framework to directly benefit from image domain pretraining used in \cite{Wu2025OmniGen2ET}. 
In terms of other components, we use Qwen2.5-VL \cite{Bai2025Qwen25VLTR} as our VLM and Flux.1 VAE \cite{Labs2025FLUX1KF} as our visual encoder-decoder.
The two MLP layers ensure channel dimension of both textual and visual features are common. These conditional inputs are simply appended to the DiT input sequence. 
Further details in \Cref{app:arch_more}. 

\subsection{Optical Flow Representation}
In contrast to prior future optical flow generation works \cite{Xu2024FlowAT,Ran25LangToMo}, we adopt an RGB space representation for optical flow (OF). This allows us to directly benefit from existing powerful VAE models to encode our OF, in contrast to training or finetuning the VAE model \cite{Xu2024FlowAT,Ran25LangToMo}.
Motivated by \cite{Chefer2025VideoJAMJA,Zhong2025FlowVLATI}, we explore representing OF in RGB formats by mapping the polar coordinate values of OF to the HSV color space. We map magnitude, rotation, and a linear combination of these two from the polar representation into the three H-S-V channels respectively. We tune the scaling values for these mappings to maximize visual continuity across frames (e.g. no flickering) and to resemble RGB images (e.g. animated graphics). See \Cref{app:pm_hsv} for further details.

\subsection{Training}
\label{subsec:train}
Our \modelname framework is trained end-to-end to predict future optical flow sequences conditioned on image sequences and paired textual captions. 
During training, the target optical flow sequence, $\vy$, is calculated using a suitable optical flow calculation algorithm, $\gF$, which can access the future frames for the calculation. 
We use $\vy[i] = \gF(\vx_{i}, \vx_{i+1})$ where $\cdot[i]$ indexes the i-th frame of each sequence.
These optical flow targets are encoded with the VAE encoder to obtain $\vf_y$ which is in turn used with a flow matching diffusion loss as the training objective. 

Let us define the learnable components of our framework, i.e. the two MLPs and DiT,  as $\gD$ such that 
$ \gD(\vf_c, \vf_v, \vn) 
\to \hat{\vf}_y $ where $\vn$ is noise input and $\hat{\vf}_y$ is predicted tensor output of the DiT module. 
Given targets $\vf_y$, corrupted targets $\tilde{\vf}_y$, and noise $\vf_0$ we obtain our training objective as, 
\begin{align}
\label{eq:fm_loss}
\mathcal{L}_{\text{FM}}(\theta) &= \lVert \{\gD(\vf_c, \vf_v, \tilde{\vf}_y) - (\vf_y - \vf_0) \rVert_2^2 \ 
\end{align}
where $\theta$ are our learnable parameters. 
During training, we perform classifier-free guidance dropping the textual condition ($\vf_c$) or the visual condition ($\vf_v$). 
We also partially mask our visual condition ($\vf_v$) along time (e.g. $\vx_{t-1}$ is masked) and viewpoint axes.
Note that our visual observation, $\vx_t$, may contain images corresponding to multiple viewpoints, hence the viewpoint masking.

\begin{figure}[t]
    \centering
    \includegraphics[width=0.95\linewidth]{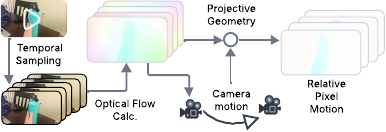}
    \vspace{-0.5em}
    \caption{
    \textbf{Relative Optical Flow Calculation:}
    We illustrate the key stages of the algorithm for calculating optical flow targets for our training.
    }
    \label{fig:pipeline}
    \vspace{-0.5em}
\end{figure}

\vspace{0.5em}
\bhdr{Relative Optical Flow Calculation:}
We construct our framework to train from internet videos of human activities (e.g. taken from a mobile phone), where each video may contain both camera (i.e. ego) motion and object motions. 
We construct our optical flow calculation algorithm, $\gF$, accounting for these possible structures of training data. 
Given a current-future frame pair, we first use an off-the-shelf optical flow calculation algorithm \cite{teed2020raft} to calculate flow vectors. Next, motivated by prior work \cite{Yang2025MagmaAF}, we use homography techniques with deep features \cite{Le2020DeepHE} to estimate the camera motion. Then we utilize the camera motion estimates with projective geometry to calculate flow vectors relative to the camera. Object motions are disentangled from camera motions in these relative optical flow vectors. We provide further details on this calculation in \Cref{app:pm_calc}. 
We also note how motion in natural videos is not uniformly distributed. To tackle this, given optical flow between frame pairs, we take the top-k percent of flow values as a proxy to motion between those frame pairs. We select frame sequences such that the motion between consecutive pairs exceeds a specific threshold. The final optical flow targets used to train our framework are calculated for these filtered frame sequences. Further details of our frame selection in \Cref{app:frame_sel}. 
Given the computational cost of our data preprocessing, the initial frame selection utilizes a fast optical calculation \cite{Lucas1981AnII} on low spatial resolution frames. Relative optical flow calculation is performed on original resolution images. Both steps are performed offline as one time processes, resulting in similar sampling costs as uniform frame sampling during training. 

\subsection{Inference}
Our VLM and VAE encoder act as feature extractors, followed by the two MLP layers projecting these features to a common dimension. 
The DiT module then uses these features as conditional inputs, and together with noise vectors performs $k$ iterations of reverse diffusion. The value $k$ is varied based on the downstream application.
For the DiT component, we perform classifier-free guidance on the text and visual features.

\subsection{Downstream Use Case I: Robot Control}
\label{subsec:robotic}

We first validate the usefulness of our \modelname framework through language driven robot control or vision-language-action (VLA). These tasks require robust motion awareness and precise language-based control. Motivated by prior work \cite{Hu2024VideoPP,Ran25LangToMo}, we integrate the future optical flow predicted by \modelname into a diffusion policy network (DPN) \cite{Reuss2024MultimodalDT}. This DPN processes the predicted future optical flow alongside textual inputs and current state information, mapping these multimodal inputs directly into action vectors that control a robotic manipulator. We provide an overview of our robot control pipeline in Figure \ref{fig:arch}.

Given the embodiment awareness necessary for robotic manipulation, we finetune our framework for downstream robotic tasks. Aligning with established methodologies \cite{Hu2024VideoPP,Ran25LangToMo}, this process comprises two stages: (1) future optical flow prediction finetuning on robotic video-caption data, and (2) action head training on robotic demonstration data. A key distinguishing feature of our work is the future optical flow finetuning phase, which explicitly accounts for the two types of viewpoints, fixed external and moving wrist cameras, inherent in robotic data. We achieve this by ensuring cross-view conditioning within the DiT backbone and augmenting our prediction targets ($\mathbf{f}_y$) to encompass optical flow from both viewpoint types. These architectural and data handling innovations differentiate our framework from previous robot control approaches \cite{Hu2024VideoPP,Ran25LangToMo}. For final robot control, the predicted motions are mapped to robotic actions with the diffusion policy network from \cite{Hu2024VideoPP} modified to condition on our future optical flow predictions.

\subsection{Downstream Use Case II: Video Generation}
\label{subsec:t2v}

We further demonstrate the versatility of \modelname framework by exploring Text-to-Video (T2V) generation focused on purely natural language-based motion control.
While prior motion-guided video generation methods typically rely on low-level visual inputs like pixel-level trajectories or motion masks generated by humans through suitable software interfaces \cite{yin2023dragnuwa,Wu2024DragAnything,Burgert2025GowiththeFlowMV}, we investigate the use of textual descriptions as a direct, human-centric motion control signal.
We establish a two-stage pipeline by connecting \modelname with the existing video synthesis model, Go-with-the-Flow (GWTF) \cite{Burgert2025GowiththeFlowMV}. First, \modelname takes a static visual observation (the initial frame) and a textual description of the desired future motion as input. Conditioned on these inputs, \modelname generates a sequence of future optical flow frames. This sequence is subsequently interpolated to create a dense motion signal. In the second stage, this dense motion and the initial frame are passed to GWTF, which synthesizes the final video sequence, ensuring the output realizes the motion pattern defined in the original text. We provide a high level overview of this extended setup in \Cref{fig:arch}.
Compared to single-stage T2V methods \cite{Yang2024CogVideoXTD,Wang2025WanOA,Kong2024HunyuanVideoAS}, our approach is computationally more intensive but offers two distinct advantages. It demonstrates superior fidelity to complex textual motion instructions, and the interpretable future optical flow sequence provides a crucial, intermediate layer of transparency to the generation process. We also highlight this as an experiment to validate the strength of language-driven future optical flow forecasting ability learned by our framework.

\begin{table*}[t]
\centering
\small
\def\arraystretch{1.1}  
\setlength\tabcolsep{1.0em}  
\scalebox{0.9}{
\begin{tabular}{lcllllll}
\toprule
\multirow{2}{*}{\textbf{Method}}& 
\multirow{2}{*}{\textbf{Training Data}} 
& \multicolumn{5}{c}{\textbf{$i^{th}$ Task Success Rate $\uparrow$}} 
& \multirow{2}{*}{\textbf{Avg. Len $\uparrow$}}
\\ \cline{3-7} 
& & \textbf{1} & \textbf{2} & \textbf{3} & \textbf{4} & \textbf{5} & \\ \midrule
RT-1 \cite{rt1}                             & 100\% ABC & 53.3 & 22.2 &  9.4 &  3.8 &  1.3 & 0.90 \\
Diffusion Policy \cite{Chi2023DiffusionPV}  & 100\% ABC & 40.2 & 12.3 &  2.6 &  0.8 &  0.0 & 0.56 \\ 
Robo-Flamingo \cite{Li2023VisionLanguageFM} & 100\% ABC & 82.4 & 61.9 & 46.6 & 33.1 & 23.5 & 2.47 \\
Uni-Pi \cite{du2023learning}                & 100\% ABC & 56.0 & 16.0 &  8.0 &  8.0 &  4.0 & 0.92 \\
MDT \cite{Reuss2024MultimodalDT}            & 100\% ABC & 63.1 & 42.9 & 24.7 & 15.1 &  9.1 & 1.55 \\
Susie \cite{Black2023ZeroShotRM}            & 100\% ABC & 87.0 & 69.0 & 49.0 & 38.0 & 26.0 & 2.69 \\
GR-1 \cite{wu2023unleashing}                & 100\% ABC & 85.4 & 71.2 & 59.6 & 49.7 & 40.1 & 3.06 \\ 
Vidman \cite{Wen2024VidManEI}               & 100\% ABC & 91.5 & 76.4 & 68.2 & 59.2 & 46.7 & 3.42 \\ 
RoboUniview \cite{Liu2024RoboUniViewVM}     & 100\% ABC & 94.2 & 84.2 & 73.4 & 62.2 & 50.7 & 3.65 \\ 
LTM \cite{Ran25LangToMo}                    & 100\% ABC & 97.1 & 82.4 & 72.8 & 67.2 & 60.6 & 3.81 \\ 
DreamVLA \cite{Zhang2025DreamVLAAV}         & 100\% ABC & 98.2 & 94.6 & 89.5 & 83.4 & 78.1 & 4.44 \\ \midrule
VPP \cite{Hu2024VideoPP}                    & 100\% ABC & 96.5 & 90.9 & 86.6 & 82.0 & 76.9 & 4.33 \\ \rowcolor{Gray}
\modelname        & 100\% ABC & 98.8\inc{2.3} & 95.0 \inc{4.1} & 90.4\inc{3.8} & 84.6\inc{2.6} & 78.7\inc{1.8} & 4.48\inc{0.15} \\ \midrule
GR-1 \cite{wu2023unleashing}                &  10\% ABC & 67.2 & 37.1 & 19.8 & 10.8 & 06.9 & 1.41 \\
VPP \cite{Hu2024VideoPP}                    &  10\% ABC & 87.8 & 74.6 & 63.2 & 54.0 & 45.3 & 3.25 \\ \rowcolor{Gray}
\modelname               &  10\% ABC & 90.4\inc{2.6} & 77.4\inc{2.8} & 65.9\inc{2.7} & 62.8\inc{8.8} & 46.9\inc{1.6} & 3.43\inc{0.18} \\
\bottomrule
\end{tabular}
}
\vspace{-0.5em}
\caption{
\textbf{CALVIN Evaluation:}
Zero-shot long-horizon evaluation on the Calvin ABC$\rightarrow$D benchmark where agent is asked to complete five chained tasks sequentially based on instructions. We report success rate (\%) for each sequential task and the average length of task completion following standard protocol. Improvements over VPP \cite{Hu2024VideoPP} baseline reported for our method.
}
\label{tbl:res_calvin}
\end{table*}

\begin{table*}[t]
\centering
\small
\def\arraystretch{1.1}  
\setlength\tabcolsep{0.8em}  
\scalebox{0.9}{
\begin{tabular}{lllllll}
    \toprule
    \textbf{Method} & 
    \textbf{Handover Block} & 
    \textbf{Handover Mic} & 
    \textbf{Pick Diverse Bottles} & 
    \textbf{Pick Dual Bottles} & 
    \textbf{Place Dual Shoes} &
    \textbf{Average} \\
    \midrule
    RDT~\cite{liu2024rdt}         & 45 & 90 &  2 & 42 &  4 & 36.6 \\
    ACT~\cite{zhao2023learning}   & 42 & 85 &  7 & 31 &  9 & 34.8 \\
    DP~\cite{Chi2023DiffusionPV}  & 10 & 53 &  6 & 24 &  8 & 20.2 \\
    DP3~\cite{Ze2024DP3}          & 70 &100 & 52 & 60 & 13 & 59.0 \\
    $\pi_{0}$~\cite{Black20240AV} & 45 & 98 & 27 & 57 & 15 & 48.4 \\ \midrule
    VPP~\cite{Hu2024VideoPP}      & 54 & 80 & 60 & 63 & 52 & 61.8 \\ \rowcolor{Gray}
    \modelname (ours)             & 61\inc{7} & 87\inc{7} & 67\inc{7} & 68\inc{5} & 60\inc{8} & 68.6\inc{6.8} \\ 
\bottomrule
\end{tabular}
}
\vspace{-0.5em}
\caption{\textbf{RoboTwin Evaluation:}
We report success rates on the RoboTwin 2.0 benchmark following their official protocol on the easy setting. This environment contains a bimanual robot. The five tasks are selected based on necessity of both arms to complete the task. 
Our proposed \modelname achieves consistent improvements over the VPP baseline implemented under identical settings.
}
\label{tbl:robotwin}
\end{table*}

\begin{table*}[t]
\centering
\small
\def\arraystretch{1.1}  
\setlength\tabcolsep{1.1em}  
\scalebox{0.9}{
\begin{tabular}{lllllll}
\toprule
Method                   & SSIM $\uparrow$&PSNR $\uparrow$&LPIPS $\downarrow$&FVD$\downarrow$&KVD$\downarrow$& 
MF $\uparrow$ \cite{yatim2024space} \\ \midrule
Seer \cite{Gu2023SeerLI}                   & 41.8          & 10.71           & 58.8          & 287.46         & 81.31          & ---    \\
Dynamicrafter \cite{xing2024dynamicrafter} & ---      & ---        & ---           & 204.11         & 31.81          & ---    \\
CosHand-I \cite{Sudhakar2024ControllingTW} & 61.5          & 16.87           & 31.3          & 91.18          & 19.24          & 0.432  \\
CosHand-A \cite{Sudhakar2024ControllingTW} & 53.1          & 14.92           & 40.8          & 90.30          & 13.68          & 0.570  \\ 
InterDyn \cite{Akkerman2024InterDynCI}     & 66.4          & 18.60           & 26.0          & 19.27          & 1.99           & 0.633  \\
InterDyn-R \cite{Akkerman2024InterDynCI}   & 68.0          & 19.04           & 25.2          & 22.22          & 2.09           & 0.641  \\ \midrule
CogVideoX \cite{Yang2024CogVideoXTD}       & 67.2          & 21.51           & 30.3          & 78.47          & 12.46          & 0.594  \\ \rowcolor{Gray}
\modelname (ours)        & 68.4\inc{1.2} & 22.26\inc{0.75} & 28.5\inc{1.8} & 75.39\inc{3.08}& 11.38\inc{1.08}& 0.662\inc{0.068}  \\ 
\bottomrule
\end{tabular}
}
\caption{
\textbf{Evaluation of Language-Driven Motion Control in T2V:}
We evaluate the ability of our T2V pipeline for language-driven motion control using the motion heavy SSv2 dataset following prior work \cite{Akkerman2024InterDynCI}. Over the CogVideoX baseline, our \modelname framework shows consistent improvements in generation quality. 
}
\label{tbl:t2v_res}
\end{table*}


\section{Experiments}
\label{sec:exp}

In this section, we first detail our experimental setup followed by evaluations on language-driven robotics manipulation and motion focused text-to-video generation. Finally we provide ablations of our \modelname framework to validate our design choices. 

\vspace{0.5em}
\bhdr{Implementation Details:}
Our \modelname framework contains three key components: VLM, VAE, and DiT. 
We use the Qwen2.5-VL \cite{Bai2025Qwen25VLTR} architecture for our VLM and the Flux.1 \cite{Labs2025FLUX1KF} architecture for our VAE. For both models, we initialize with their official pretrained weights. 
Both the VLM and VAE weights are kept frozen. 
Our DiT uses a standard diffusion transformer architecture following OmniGen \cite{Wu2025OmniGen2ET} with modifications for video generation as described in \Cref{subsec:arch}.
We initialize the DiT (including projector layer MLP) with weights from a model pretrained on image editing tasks from \cite{Wu2025OmniGen2ET}. 

\vspace{0.5em}
\bhdr{Training Details:}
We train our \modelname framework on the human activity data from the train split of Something-Something-V2 dataset \cite{Goyal2017TheS} and the EgoDex dataset \cite{Hoque2025EgoDexLD}. Our training data contains roughly 500,000 video-caption pairs.  
We train our model on 8xH200 GPUs with a global batch size of 4096 for 100 epochs. We note how different frame subsequences are sampled from the same video across epochs based on motion based sampling as explained in \Cref{subsec:train}. 
For robotic manipulation, we additionally finetune our model on all available video data (without access to action trajectory labels) on the downstream tasks following prior robotic manipulation works \cite{Hu2024VideoPP,Zhang2025DreamVLAAV,Ran25LangToMo}. 

\subsection{Language-Driven Robot Manipulation}
We evaluate on two robot manipulation benchmarks, CALVIN \cite{Mees2021CALVINAB} and RoboTwin 2.0 \cite{chen2025robotwin}. 
The CALVIN benchmark is an open-source, simulated environment designed for long-horizon, language-conditioned tasks. It challenges agents to perform complex, multi-step manipulation sequences using continuous control, with task goal signals specified only via unconstrained natural language instructions. This dataset focuses on developing policies that can generalize to novel language instructions and unseen environments, making it an ideal evaluation task for our framework.
For this task, we utilize the CALVIN ABC$\rightarrow$D setting, where training data is available only for ABC environments, while evaluation is performed zero-shot on the novel D environment. 
Given the multi-step sequential manipulation tasks within this benchmark, the standard evaluation metrics are $i^{th}$ task success rate and the average length of task stages completed. We report this standard metrics from the benchmark used across prior work \cite{Hu2024VideoPP,Ran25LangToMo} for our evaluations.  
The RoboTwin 2.0 framework provides a scalable approach for bimanual (dual-arm) robotic manipulation, specifically addressing the limitations of oversimplified simulations for two-arm tasks. We select 5 tasks that specifically require utilizing both arms for manipulation in order to successfully complete the task. This benchmark allows evaluating the generality of our framework across both single and dual arm robotic environments. The standard evaluation metric for this benchmark is success rate which we use in our evaluations, along with the average success rate across all tasks.

\bhdr{CALVIN Results:}
Our results reported in \Cref{tbl:res_calvin} demonstrate state-of-the-art performance of our \modelname framework in long-horizon, zero-shot robot manipulation. 
All evaluations follow standard protocol \cite{Mees2021CALVINAB}. Results for baselines are reported directly from prior work \cite{Hu2024VideoPP,Zhang2025DreamVLAAV}.  
Training on the full split for the ABC$\rightarrow$D setting, our \modelname framework registers the highest success rate across all five chained tasks, culminating in a Task 5 Success Rate of $0.787$ and the highest Average Length of task completion at $4.48$, marginally surpassing the prior best model, DreamVLA ($4.44$). This strong performance confirms our method's superior ability to generalize to novel language instructions for sequential, multi-step manipulation. Furthermore, in the data-limited regime (using only $10\%$ of the ABC dataset), our method continues to excel, achieving an average length of $3.43$, which is noticeably higher than other models in this setting, highlighting its data efficiency. Similar trends of data efficiency have been observed in prior works utilizing motion representations \cite{Ran25LangToMo,Gao2017Im2FlowMH}, another benefit of our framework. 

\bhdr{RoboTwin Results:}
We report RoboTwin results in \Cref{tbl:robotwin}, demonstrating the effectiveness of our \modelname framework in complex bimanual manipulation tasks. 
Results for baselines are directly from the official leaderboard \cite{chen2025robotwin}, with attempted replication following the official codebase. 
We note that we were unable to replicate the baseline results reported in leaderboard for some tasks (e.g. Handover Mic), but nevertheless report official values from the benchmark leaderboard.  
We replicate the VPP baseline \cite{Hu2024VideoPP} following their official codebase training under identical settings as our \modelname framework, enabling a direct and fair point of comparison.
On our selected task subset, our model achieves an average success rate of $68.6\%$, significantly outperforming our baseline model, VPP ($61.8\%$), which is trained similarly on web human videos, but with frame prediction instead of motion prediction. Our \modelname outperforms this baseline consistently across all tasks, demonstrating the strength of motion prediction over frame prediction. 
In comparison to other baselines, our model notably shows robust performance across tasks (e.g. Place Dual Shoes), highlighting its better generalization ability. 
We take these results as confirmation on the utility of \modelname framework for solving complex robotic tasks requiring cooperative bimanual manipulation.

\subsection{Motion Control in Video Generation}
Our second downstream task focuses on text-to-video generation with focus on language-driven motion control aspects. 
\Cref{tbl:t2v_res} presents results for our evaluation on the validation split of SSv2 dataset following prior work \cite{Akkerman2024InterDynCI}. Our evaluation protocol is identical to \cite{Akkerman2024InterDynCI} with results for baselines \cite{Gu2023SeerLI,xing2024dynamicrafter,Sudhakar2024ControllingTW,Akkerman2024InterDynCI} reported directly from \cite{Akkerman2024InterDynCI}. We implement and evaluate the CogVideoX baseline \cite{Yang2024CogVideoXTD} under identical settings as our \modelname framework. We achieve consistent improvements over the CogVideoX baseline across all major metrics. Notably, while most prior controllable-generation baselines (e.g., \cite{Sudhakar2024ControllingTW,Akkerman2024InterDynCI}) rely on auxiliary control signals such as hand or object masks over time—inputs that are not freely available at inference, our framework operates solely from language prompts and visual context. Despite this weaker supervision at test-time, \modelname attains comparable or superior performance, underscoring its capacity to learn text-conditioned motion representations that generalize without explicit spatiotemporal guidance. 
We take these results as indication to how motion-aware language grounding alone can yield coherent, controllable dynamics in text-to-video generation.

\subsection{Ablations}

\begin{table}[t]
\centering
\small
\def\arraystretch{1.2}  
\setlength\tabcolsep{1.2em}  
\scalebox{0.86}{
\begin{tabular}{lcl}
\toprule
\begin{tabular}[c]{@{}l@{}}Pretrain \\ Dataset\end{tabular} & Train Steps & Avg Len \\ \midrule
DROID     & 2000   & 4.04    \\ \rowcolor{Gray}
SSv2      & 2000   & 4.39\inc{0.35}    \\ \bottomrule
\end{tabular}
}
\vspace{-0.5em}
\caption{\textbf{Ablation on Human Web Video Pretraining:}
The abundance of web videos of human activities and their inexpensive collection in comparison to teleoperated robotic manipulation videos allows training with significantly more human activity data. For fair investigation on impact of human videos, we pretrain our framework for a common step count on both DROID \cite{Khazatsky2024DROIDAL} and SSv2 dataset. DROID is considered a large-scale ``In-the-Wild" robotic manipulation dataset that contains diversity across scenes and actions in its teleoperated demonstration videos. 
}
\label{tbl:human_vid_ablate}
\end{table}
\begin{table}[t]
\centering
\small
\def\arraystretch{1.2}  
\setlength\tabcolsep{1.2em}  
\scalebox{0.86}{
\begin{tabular}{lcl}
\toprule
Backbone       & IE PT    & Avg Len \\ \midrule
Diffusion Only &  \xmark  & 4.01\dec{0.38}    \\ 
VLM-Diffusion  &  \xmark  & 4.14\dec{0.25}    \\ \rowcolor{Gray}
VLM-Diffusion  &  \cmark  & 4.39    \\ \bottomrule
\end{tabular}
}
\vspace{-0.5em}
\caption{\textbf{Ablation on Backbone Choice:}
We evaluate our VLM-Diffusion architecture against a diffusion only backbone following \cite{Ran25LangToMo} and demonstrate the gains in terms of learning information from web human videos that is useful for robotic tasks. We also ablate the image-editing pretraining we adopt from \cite{Wu2025OmniGen2ET} to demonstrate its utility for motion guidance prediction.
}
\label{tbl:ablate_backbone}
\end{table}

Next we analyze contributions of each design choice. For all experiments in this section, we use only the SSv2 dataset as the pretraining source for faster experimentation. This dataset provides a broad range of human activity videos with rich motion diversity, making it a suitable testbed for controlled comparisons. For evaluation, we follow the standard CALVIN setup \cite{Mees2021CALVINAB,Hu2024VideoPP}, where each evaluation sequence consists of five chained tasks. While our main experiments utilize the full 1000-sequence benchmark to ensure comprehensive assessment, the ablations are performed on a reduced subset of 400 sequences to facilitate efficient experimentation without compromising representativeness. The following analyses dissect three key factors: pretraining, backbone architecture, and motion disentangling strategy. We evaluate each for its impact on motion guidance prediction through downstream task performance.

\noindent
\textbf{Pretraining Ablation:}
\Cref{tbl:human_vid_ablate} investigates the effect of pretraining on large-scale web videos of human activities compared to teleoperated robotic demonstrations by using the full DROID dataset \cite{Khazatsky2024DROIDAL}. While DROID provides diverse in-the-wild robot manipulation demonstrations, it remains limited in coverage and motion variability relative to human activity videos such as SSv2. When pretrained for the same number of steps, our model exhibits a noticeable performance gain ($+0.35$) when trained on SSv2, confirming that web-scale human motion data significantly improves the model’s ability to infer language-conditioned motion patterns. This highlights the value of human web videos as a scalable supervision source for motion understanding, even though they are noisier and less structured than robot demonstrations. The result supports our hypothesis that large, diverse human video corpora can serve as an effective pretraining substrate for motion guidance learning prior to domain-specific adaptation.

\begin{table}[t]
\centering
\small
\def\arraystretch{1.2}  
\setlength\tabcolsep{1.2em}  
\scalebox{0.86}{
\begin{tabular}{ll}
\toprule
Method                      & Avg Len \\ \midrule 
No Video Pretraining        & 3.91\dec{0.48} \\
Static Frame Targets        & 4.28\dec{0.11} \\ 
Raw Motion Targets          & 3.89\dec{0.50} \\ \rowcolor{Gray} 
Disentangled Motion Targets & 4.39           \\ 
\bottomrule
\end{tabular}
}
\vspace{-0.5em}
\caption{\textbf{Ablation on Motion Disentangling Strategy:}
We examine the importance of motion disentangling for learning language driven dense pixel motion prediction from web videos of human activities. In line with prior work \cite{Yang2025MagmaAF}, our results demonstrate the difficulty of learning any meaningful signals without suitable disentangling of motion targets using our algorithm. 
In Row 1, we provide a baseline with no video pretraining, followed by pretraining with static frame targets (similar to VPP \cite{Hu2024VideoPP}), raw motion targets (no camera-object motion separation), and our disentangled motion targets (default) in Rows 2-4 respectively.  
}
\label{tbl:mo_alg_ablate}
\end{table}

\noindent
\textbf{Backbone Ablation:}
\Cref{tbl:ablate_backbone} explores contribution of the unified VLM–diffusion backbone relative to a diffusion-only architecture as well as the fine-grained image-editing pretraining (IE-PT) that is enabled by this VLM-diffusion architecture \cite{Wu2025OmniGen2ET}. 
The diffusion-only baseline achieves lower performance, indicating its ability to model low-level motion but limited understanding of linguistic context. Incorporating a VLM–diffusion backbone without image-editing pretraining improves performance, as the integrated vision–language transformer model (VLM) introduces stronger multimodal reasoning and better semantic grounding. The full VLM–diffusion model with image-editing pretraining achieves the highest score, confirming that pretraining on language-driven image transformation tasks facilitates cross-modal alignment between linguistic commands and visual motion cues. Together, these results validate our architectural design: combining VLM and diffusion components benefits generalizable learning of motion guidance prediction from weakly aligned web captions.

\noindent
\textbf{Ablation on Motion Target Calculation:}
Table \ref{tbl:mo_alg_ablate} examines how different formulations of motion supervision affect the ability to learn meaningful language-conditioned motion guidance prediction. Removing disentangling and using raw optical flow (Row 3) results in a significant performance drop, suggesting that camera-induced motion severely hinders the learning of object-centric motion guidance. Similarly, static frame targets (Row 2), which ignore motion altogether, lead to underfitting of temporal dynamics. In contrast, our disentangled motion targets (Row 4, default) yield the best results, demonstrating that separating object motion from camera motion provides clean, physically meaningful supervision that aligns with linguistic intent. This confirms that our motion disentangling algorithm is critical for grounding the model’s predictions in genuine object dynamics rather than pixel changes caused by viewpoint shifts.

\section{Conclusion}

We presented \modelname, a new framework for bridging language and motion through dense, pixel-level displacement prediction.
Our disentangling algorithm refines noisy web video supervision by separating object and camera motion, yielding cleaner learning signals.
The unified VLM–diffusion backbone further enables robust multimodal learning from diverse and noisy caption data.
Through extensive experiments across robotics and video generation, we demonstrate that motion-guided representations substantially enhance language-conditioned control and synthesis.
We hope this work paves the way toward generalizable, motion-aware world models that understand and act through dynamic visual grounding.

\subsection{Limitations and Future Work}
\label{subsec:limit_future}
Our \modelname model currently contains several limitations. 
It is sensitive to the text prompt (i.e. slightly different text prompts could lead to wrong predictions). For example, changing a text ``moving from right to left''  $\rightarrow$ ``moving left'' results in wrong predictions for some samples. 
In addition, the current \modelname is a relatively large model ($\sim$7B parameters in total) that is expensive to deploy in a real-time setting and requires considerable compute (at least 24GB of GPU) for inference. Further discussion in \Cref{sec:app_lim}.

On the other hand, our model captures diversity of motion patterns quite well and generates meaningful future optical flow in as less as 1 reverse diffusion (i.e. denoising) iteration. 
Across different seeds for the same frame-caption pair, our model generates a diverse distribution of mostly meaningful future optical flow: we attribute this strength to the diffusion based training. 
On fast convergence (during reverse diffusion), we note how our model is not distilled or trained specifically for this purpose; RGB image generation with identical architecture and training pipeline usually requires at least 20 sampling steps for meaningful generation.  
We hypothesize that optical flow is a less complex distribution compared to natural images (i.e. lies on a lower dimensional manifold and contains less high frequency information such as textures or patterns): this allows even a single sampling step of reverse diffusion to generate meaningful outputs. 

In future work, we plan to explore training data augmentation with automated text-label re-phrasing and model distillation into light-weight architectures to address our key limitations. We also aim to further investigate and quantify the diversity aspect of \modelname, analyze the fast convergence aspect, and explore possibility of real-time inference.

\section*{Reproducibility Statement}
To ensure the reproducibility of our results, we have made our code, trained model weights, and data configurations publicly available (see \url{fofpred.github.io}). Our method builds upon open-source pretrained models retrieved from the Hugging Face Hub. Detailed implementation specifics, hyperparameter settings, and experimental protocols are thoroughly documented in the main text and the Appendix.

\section*{Acknowledgments}
We would like to thank the broader research team at Salesforce for support, feedback and guidance. 
In particular, we thank 
Shelby Heinecke, Juntao Tan, Srinath Meadusani, Eric Xu, Matthew Fernandez, Jim Jagielski, Joyce Zheng, and Jinxuan Xu 
for technical feedback and support; 
Jeanette Berberich and Jennifer McCallion for legal coordination; 
Janna Remperas, Ian Thomas, Mitra Mitchell, and Alex Dalton 
for logistical help and encouragement throughout the project.




{
    \small
    \bibliographystyle{ieeenat_fullname}
    \bibliography{main}
}

\clearpage
\newpage

\maketitlesupplementary
\appendix

\section*{Contributions}
KR led the project by implementing the preliminary idea of language-driven OF prediction, building the codebase for experimentation, and performing most of the evaluations.
HZ discussed all aspects of the project, contributed to several key design choices (on model architecture, OF calculation, video evaluations), and helped debug several technical issues.  
YF discussed multiple aspects of the robotic evaluation pipelines, helped implement the RoboTwin evaluations, and performed several robotic baseline evaluations.   
LY contributed to building the video evaluation pipeline, helped investigate our novelty against prior works, and discussed several aspects of our video generation pipeline. 
LX proposed using the unified VLM-Diffusion architecture, provided feedback on our data and training pipeline implementations, and discussed several early results. 

RX helped streamline early exploration of the idea, supported scaling up our training pipeline, and provided several critical feedback on project design. CX \& SS provided the strategic vision for the project, oversaw the research environment, and shaped the high-level framing of the research problem.
MR set the direction for robotic downstream tasks and discussed multiple aspects of the project idea, scope, and implementation.
JN organized the overall project, set the research direction, and discussed all aspects of the project idea, scope, and implementation.

\renewcommand{\thetable}{A.\arabic{table}}
\renewcommand{\thefigure}{A.\arabic{figure}}
\setcounter{table}{0}
\setcounter{figure}{0}

\vspace{1.0em}

\addtocontents{toc}{\protect\setcounter{tocdepth}{1}}
\etocsettocstyle{\section*{Appendix Contents}}{}

\renewcommand{\cftsecaftersnum}{} 
\setlength{\cftsecnumwidth}{1.7em} 

\hypersetup{linkcolor=black}        
\tableofcontents
\hypersetup{linkcolor=blue}         

\vspace{1.5em}

\section{Additional Architectural Details}
\label{app:arch_more}

We provide additional details of our architecture in this section, focusing on the conditional input processing and the modifications made to the Diffusion Transformer (DiT).
To recap, our overall architecture contains 3 main components: the VLM (3B parameters), VAE (83M parameters), and DiT (4B parameters). The DiT (diffusion transformer) module is the key component that we modify and train to construct our \modelname model. 

\subsection{Conditional Input Processing}
The core architecture utilizes two distinct conditional features: $\vf_c$ (textual) and $\vf_v$ (visual), which are passed through Multi-Layer Perceptrons (MLPs) to ensure dimensional compatibility with the Diffusion Transformer.

\begin{itemize}[leftmargin=1.5em,noitemsep,topsep=0.5ex,itemsep=0.0ex,partopsep=0ex,parsep=1ex]
    \item \textbf{Textual Feature Projection:} The textual feature $\vf_c$, obtained from the Qwen2.5-VL \cite{Bai2025Qwen25VLTR} Vision-Language Model (VLM), has an initial channel dimension of $2520$. This feature is projected via an MLP layer to a common channel dimension $D$. The VLM input consists of the natural language caption $c$ paired with the interleaved visual inputs $\vx_{t-1}$ and $\vx_t$.
    
    \item \textbf{Visual Feature Projection:} The visual feature $\vf_v$, obtained from the Flux.1 VAE \cite{Labs2025FLUX1KF} encoder, has an initial channel dimension of $16$. This feature is reshaped using $2 \times 2$ grid to obtain $64 = 16 \times 2 \times 2$ channel dimension vectors, followed by projected via a different MLP layer to the same common channel dimension $D$.

    \item \textbf{Initial Noise Vector:} The noise vector, sampled from the VAE latent space, is also projected into the DiT channel dimension using the same reshaping operation and visual feature project MLP layer. 

\end{itemize}

We enforce the condition that the output dimensions of the two MLPs are equal to $D = 2520$ which is the DiT input channel dimension.
These projected conditional features, $\hat{\vf}_c$ and $\hat{\vf}_v$, are simply appended to the input noise sequence of the DiT, providing comprehensive context for the diffusion process.

\subsection{DiT Modifications for Temporal Modeling}

Our Diffusion Transformer is based on the OmniGen architecture \cite{Wu2025OmniGen2ET}, adapted to explicitly handle the temporal nature of our sequence prediction task (future optical flow $\hat{\vy}_{t+1:t+4}$ from inputs $\vx_{t-1}, \vx_t$).

\vspace{0.5em}
\bhdr{Time-Aware 3D RoPE:}
We implemented a modified RoPE module to handle the required temporal dimension within the sequence. The original OmniGen RoPE provides 3-axis position encoding $(L, H, W)$ for text length ($L$), image height ($H$), and image width ($W$). Our modification interprets these three axes as:
\begin{itemize}[leftmargin=1.5em,noitemsep,topsep=0.5ex,itemsep=0.0ex,partopsep=0ex,parsep=1ex]
    \item \textbf{Axis 1 (Text/Image Shift):} Used to assign unique base position IDs for the text tokens and for each separate image (or image sequence). This axis is now leveraged to encode the \textbf{temporal offset} or frame index for the input sequences ($\vx_{t-1}, \vx_t$) and the latent noisy output $\vy$.
    \item \textbf{Axis 2 and 3 (Image $H, W$):} Encode the spatial positions within each frame's patch grid.
\end{itemize}

Specifically, for multiple frames (e.g., $\vx_{t-1}$ and $\vx_t$ representations), the position shift variable is incremented after processing each frame. This ensures that while text tokens and $\vx_{t-1}$ tokens receive their respective base position IDs, the tokens corresponding to $\vx_{t}$ receive a unique, subsequent base position ID, which acts as a temporal index.
Overall, this ensures that the tokens are timestamped with temporal position using RoPE, allowing the transformer to distinguish between and model the relationship across the consecutive time steps in both the input and output frame sequences.

\vspace{0.5em}
\bhdr{Spatio-Temporal Attention:}
The DiT transformer blocks are updated to perform \textit{full spatio-temporal attention} over the output tokens. This means the self-attention mechanism is configured to operate over the entire input sequence $\vf_{in} = [\hat{\vf}_c, \hat{\vf}_v, \vf_y]$, where $\vf_y$ is the latent sequence of the noisy optical flow. 
This design choice allows the DiT to explicitly capture the motion dynamics and sequence dependencies required for accurate predictions of future optical flow frame sequences.

\section{Optical Flow Representation}
\label{app:pm_hsv}

The visualization algorithm begins by converting the Cartesian optical flow field, denoted as $F \in \mathbb{R}^{2 \times H \times W}$ with components $f_x$ and $f_y$, into a polar representation. The magnitude of the flow is computed as $M = \sqrt{f_x^2 + f_y^2}$ and normalized by a scaling factor $\eta = 64.0$. This normalized magnitude, $\hat{M}$, is clamped to the range $[0, 1]$. The directional angle is derived using the four-quadrant inverse tangent function, $\phi = \operatorname{arctan2}(f_y, f_x)$, and is shifted by $\pi$ to ensure the final angle $\theta$ lies within the interval $[0, 2\pi]$.

Subsequently, these polar coordinates are mapped to the HSV color space to generate an RGB image. The hue channel $H$ is assigned the flow angle $\theta$, encoding the direction of motion, while the saturation channel $S$ is defined by the normalized magnitude $\hat{M}$. The value channel $V$ is set to a constant maximum intensity of $1$ (formulated in the algorithm as $\hat{M} + (1 - \hat{M})$) - this eliminates large color variances across consecutive frames due to outliers. Finally, the resulting HSV tensor is converted into RGB format using a differentiable color space transformation.

\newcommand{\Flow}{\mathbf{F}}
\newcommand{\Point}{\mathbf{p}}
\newcommand{\Hmat}{\mathbf{H}}
\newcommand{\Grid}{\mathcal{G}}
\newcommand{\Sampled}{\mathcal{S}}

\begin{algorithm}[t]
\caption{Relative Optical Flow Calculation}
\label{alg:flow_compensation}
\begin{algorithmic}[1]
\Require
    \State $\Flow_{\text{raw}} \in \mathbb{R}^{B \times 2 \times H \times W}$: Default optical flow tensors
    \State $\tau_{\text{ransac}}$: RANSAC reprojection threshold (default 5.0)
    \State $s$: Sampling stride (default 8)
    \State $\tau_{\text{noise}}$: Post-compensation noise threshold (default 0.5)
\Ensure
    \State $\Flow_{\text{comp}}$: Camera-motion compensated flow tensors

\Statex
\Procedure{CompensateFlow}{$\Flow_{\text{raw}}, s, \tau_{\text{ransac}}, \tau_{\text{noise}}$}
    \State $B, C, H, W \gets \text{shape}(\Flow_{\text{raw}})$
    \State $\Flow_{\text{comp}} \gets \text{List}()$

    \State $\Grid \gets \{(x, y) \mid 0 \le x < W, 0 \le y < H\}$ 
    \State $\Sampled \gets \text{Sample}(\Grid, s)$; \quad $\Point_0 \gets \text{Coords}(\Sampled)$ 

    \For{$i \gets 0$ \textbf{to} $B-1$}
        \State $\mathbf{f}_i \gets \Flow_{\text{raw}}[i]$

        \Statex \quad \textit{Derive Correspondences}
        \State $\mathbf{v} \gets \text{Extract}(\mathbf{f}_i, \Sampled)$ 
        \State $\Point_1 \gets \Point_0 + \mathbf{v}$ 

        \Statex \quad \textit{Estimate Homography}
        \State $\Hmat \gets \text{FindHomography}(\Point_0, \Point_1, \text{RANSAC}, \tau_{\text{ransac}})$

        \If{$\Hmat$ is valid}
            \Statex \qquad \textit{Compute Camera Flow \& Compensation}
            \State $\Grid' \gets \text{PerspectiveTransform}(\Grid, \Hmat)$
            \State $\mathbf{f}_{\text{cam}} \gets \Grid' - \Grid$
            \State $\mathbf{f}_{\text{obj}} \gets \mathbf{f}_i - \mathbf{f}_{\text{cam}}$

            \Statex \qquad \textit{Post-Compensation Thresholding}
            \If{Threshold Enabled}
                \State $\mathbf{M} \gets \|\mathbf{f}_{\text{obj}}\|_2$
                \State $\mathbf{f}_{\text{obj}}[\mathbf{M} < \tau_{\text{noise}}] \gets 0$
            \EndIf
            \State $\mathbf{f}_{\text{final}} \gets \mathbf{f}_{\text{obj}}$
        \Else
            \State $\mathbf{f}_{\text{final}} \gets \mathbf{f}_i$ 
        \EndIf

        \State Append $\mathbf{f}_{\text{final}}$ to $\Flow_{\text{comp}}$
    \EndFor

    \State \Return $\text{Stack}(\Flow_{\text{comp}})$
\EndProcedure
\end{algorithmic}
\end{algorithm}

\section{Optical Flow Calculation}
\label{app:pm_calc}

The proposed method (Algorithm~\ref{alg:flow_compensation}) compensates for camera motion within dense optical flow fields by leveraging the flow data itself to derive dense correspondences. Rather than relying on computationally expensive sparse feature extraction, we uniformly sample grid points from the source frame and project them into the target frame using the raw flow vectors $\Flow_{\text{raw}}$. These calculated point pairs serve as inputs for a RANSAC-based homography estimation, which computes a transformation matrix $\mathbf{H}$ representing the global camera motion. A dense ``camera flow'' field, $\Flow_{\text{cam}}$, is subsequently synthesized by applying $\mathbf{H}$ to the entire coordinate grid and calculating the displacement vectors. Finally, the object-centric motion is isolated by subtracting the estimated camera flow from the raw flow, such that $\Flow_{\text{obj}} = \Flow_{\text{raw}} - \Flow_{\text{cam}}$. A post-processing magnitude threshold is applied to $\Flow_{\text{obj}}$ to suppress residual noise and artifacts arising from imperfect alignment.

We reiterate that this one-time process is run offline. For our 500,000 training videos, we are able to complete processing using 4 A100 GPUs in roughly 30 hours. 
This timing also includes the motion-guided frame sampling (negligible time for this operation compared to relative flow calculation) that we describe in the next section. 

\section{Motion-Guided Frame Sampling}
\label{app:frame_sel}

\begin{algorithm}[t]
\caption{Motion-Aware Frame Selection}
\label{algo:frame_selection}
\begin{algorithmic}[1]
\Require Video Sequence $V = \{I_1, I_2, \dots, I_N\}$, Threshold $\tau$, Top-Percentile $k$
\Ensure Selected Frame Indices $S$
\State Initialize selected set $S \leftarrow \emptyset$
\For{$t \leftarrow 1$ \textbf{to} $N-1$}
    \State \Comment{Step 1: Spatial Downsampling for efficiency}
    \State $I^{low}_t, I^{low}_{t+1} \leftarrow \text{Resize}(I_t, I_{t+1}, 32 \times 32)$
    
    \State \Comment{Step 2: Fast Optical Flow Approximation}
    \State $F_{LK} \leftarrow \text{LucasKanade}(I^{low}_t, I^{low}_{t+1})$
    
    \State \Comment{Step 3: Compute Motion Proxy}
    \State $M \leftarrow \| F_{LK} \|_2$ \Comment{Calculate L2 norm of flow vectors}
    \State $\mu_{proxy} \leftarrow \text{Percentile}(M, k)$ \Comment{Extract top-k\% value}
    
    \State \Comment{Step 4: Threshold Filtering}
    \If{$\mu_{proxy} > \tau$}
        \State $S \leftarrow S \cup \{t\}$
    \EndIf
\EndFor
\State \Return $S$
\end{algorithmic}
\end{algorithm}

Training video prediction models on natural sequences requires careful data curation, as large portions of raw video may contain static scenes or imperceptible motion that contribute little to the learning process. To address this, we employ a two-stage filtering pipeline, detailed in Algorithm~\ref{algo:frame_selection}, which acts as a computational gate to prioritize high-motion segments before generating expensive ground-truth labels.
Calculating dense optical flow (e.g., using RAFT or PWC-Net) for every frame pair in a large-scale dataset is computationally prohibitive. To circumvent this bottleneck, we utilize a lightweight approximation strategy (Lines 4--6). We first spatially downsample all input frames to a resolution of $32 \times 32$ pixels. This reduction removes high-frequency textures and compression artifacts while preserving dominant structural motion. On these low-resolution pairs ($I^{low}$), we apply the Lucas-Kanade method \cite{Lucas1981AnII}. Unlike deep learning approaches, Lucas-Kanade relies on local least-squares optimization, which converges rapidly on small spatial grids ($32 \times 32$), allowing for high-throughput processing of millions of frames.
To robustly distinguish meaningful scene activity from background noise, we derive a scalar motion proxy, $\mu_{proxy}$, from the raw flow field (Lines 8--9). We compute the magnitude (L2 norm) of the flow vectors and select the top-$k$ percentile value (k = 10) rather than the mean. This percentile-based approach ensures that the metric is driven by the moving objects within the scene, rather than being diluted by large static background regions. Frame pairs are included in the final training dataset only if this proxy value exceeds an empirical threshold $\tau$ (Lines 11--12), ensuring the model focuses on sequences with significant temporal dynamics.

The result of this motion-based frame filtering is the elimination of seemingly static regions of the video. In practice, we observe that certain static regions are removed while most frames in other motion-heavy regions remains. The resulting filtered videos still produce coherent video segments without breaking the natural temporal continuity of the motion. The low threshold for motion filtering (we use a $5$ pixel length as threshold for $256 \times 256$ images) is crucial for this. We set this based on empirical observation over a randomly selected set of videos. We highlight again that tune the filtering process is important to ensure coherent frame sequences, instead of generating frame sequences that may contain large discontinuities.

\section{Additional Ablations}
\label{app:add_abla}


We provide two new ablations on using dense optical flow as our representation and the significance of the motion signal passed into our robot action policy in \Cref{app:tbl_dense_ablate} and \Cref{app:tbl_motion_ablate} respectively. 

\vspace{0.5em}
\bhdr{Dense vs. Sparse Motion:}
We investigate the impact of dense vs. sparse motion representations on the downstream robot control task in \Cref{app:tbl_dense_ablate}, reporting average length on the CALVIN benchmark. Training and evaluation settings are identical to our ablations in the main paper. 

To isolate the value of our dense representation, we compare our method against two sparse variants: a naive baseline where our predicted future optical flows are sub-sampled to a $16 \times 16$ grid, and a re-implementation of the sparse trajectory model, ATM \cite{Wen2023AnypointTM}, trained on our dataset. The results demonstrate a distinct advantage for spatially dense motion information. Our default dense model achieves an average length of 4.39, significantly outperforming the ATM baseline (2.92) and the sub-sampled variant (1.24). This suggests that the fine-grained, pixel-level dynamics captured by dense optical flow are essential for precise robotic manipulation, whereas sparse representations fail to capture the complete motion context required for complex tasks.

\vspace{0.5em}
\bhdr{Ablations on Motion Forecast Conditioning:} 
We investigate the usefulness of motion forecasting as opposed to generic visual representations for our downstream robot control task in \Cref{app:tbl_motion_ablate}. We compare our full framework against two baselines: one where motion input is entirely removed (note that our VLM input is a representation of both images and text), and another where the motion input is replaced by static visual embeddings from our VAE encoder. The results are decisive; the policy fails almost entirely without motion input (0.02) and shows only marginal improvement when provided with the static visual features (0.52). 
In contrast, conditioning on our predicted future optical flow embeddings yields a score of 4.39. This confirms that the predicted optical flow provides unique, critical dynamic information that static visual representations cannot substitute.

\begin{table}[t]
\centering
\small
\def\arraystretch{1.2}  
\setlength\tabcolsep{1.2em}  
\scalebox{0.86}{
\begin{tabular}{lcl}
\toprule
Method          & Motion & Avg Len $\uparrow$  \\ \midrule 
Naive Baseline            & Sparse & 1.24\dec{3.15}  \\
ATM             & Sparse & 2.92\dec{1.47}  \\ \rowcolor{Gray}
Ours (default)  & Dense  & 4.39    \\
\bottomrule  
\end{tabular}}
\caption{\textbf{Dense vs. Sparse Motion:}
We report the average length metric (Avg Len) on CALVIN benchmark.
We demonstrate the significance of our dense motion representation. 
The first row uses a naive sub-sampling of our predicted future optical flows to obtain motion for points on a uniform $16 \times 16$ grid on the image. This variant uses training and inference identical to ours. 
The second row re-implements the ATM \cite{Wen2023AnypointTM} approach with training on our same data but inference similar to the original work. 
We highlight the clear improvements arising from our proposed method. 
}
\label{app:tbl_dense_ablate}
\end{table}

\begin{table}[t]
\centering
\small
\def\arraystretch{1.2}  
\setlength\tabcolsep{1.2em}  
\scalebox{0.86}{
\begin{tabular}{lcl}
\toprule
Method             & Motion Input & Avg Len $\uparrow$  \\ \midrule
No motion input    & \xmark       & 0.02\dec{4.37}    \\
Static visual input & \xmark       & 0.52\dec{3.87}    \\ \rowcolor{Gray}
Ours (default)     & \cmark       & 4.39    \\ 
\bottomrule
\end{tabular}
}
\caption{\textbf{Motion Input Ablation:}
We report the average length metric (Avg Len) on CALVIN benchmark.
For our robot control extension, we ablate the motion input under two settings. First (row 1) we remove the motion input leaving only the state and text goal inputs. Next (row 2) we replace the motion input with an embedding (from our VAE) of the visual input. Our motion inputs in the form of future optical flow clearly leads to a performance improvement. 
}
\label{app:tbl_motion_ablate}
\end{table}

\section{Detailed Limitations}
\label{sec:app_lim}

We have discussed limitations and future directions of our work in Section~\ref{subsec:limit_future} of the main paper. In this section, we expand our discussion on limitation of our proposed \modelname model in detail, particularly focused on our base model (i.e. the result of our large-scale pretraining). 

\begin{figure}
    \centering
    \includegraphics[width=\linewidth]{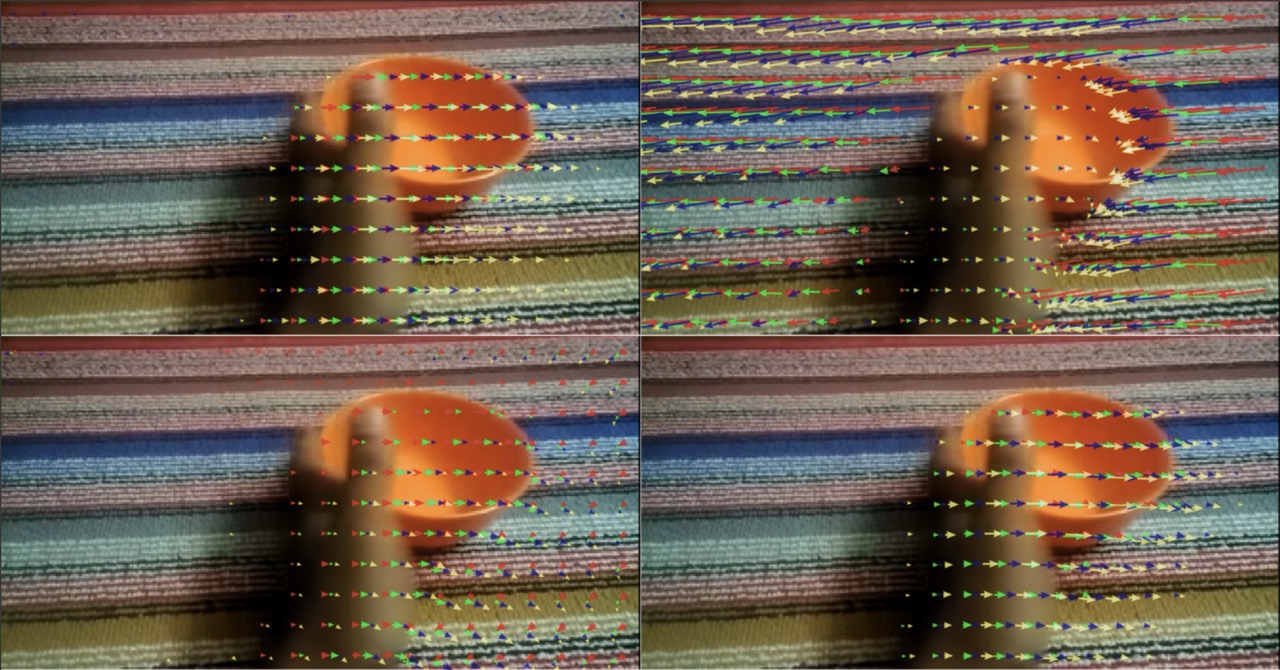}    
    \caption{\textbf{Sensitivity to seed:}
    We visualize 4 \modelname predictions for the same image and prompt, \texttt{"Moving the bowl from left to right"}, but using 4 different starting noise vectors for the reverse diffusion process. 
    Notice how the upper-right corner conflates the object motion with a camera motion instead; however this camera motion does correspond to the object motion described in the provided prompt.
    In the lower left image, in addition to the desired object motion, we again observe a slight amount of corresponding camera motion.  
    }
    \label{fig:app_lim2}
\end{figure}

In \Cref{fig:app_lim2}, we explore the diversity of our \modelname base model predictions. Across different seeds, we notice unexpected camera motion sometimes. While often relevant to the provides motion prompt, this does not explicit capture the object motions that we desire from our model. 

\begin{figure*}
    \centering
    \includegraphics[width=\linewidth]{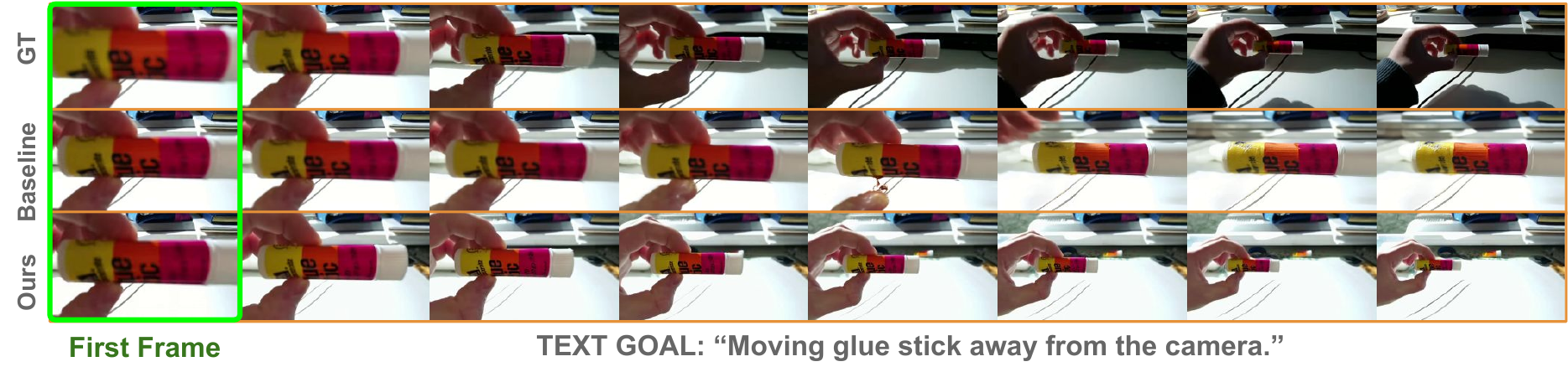}
    \includegraphics[width=\linewidth]{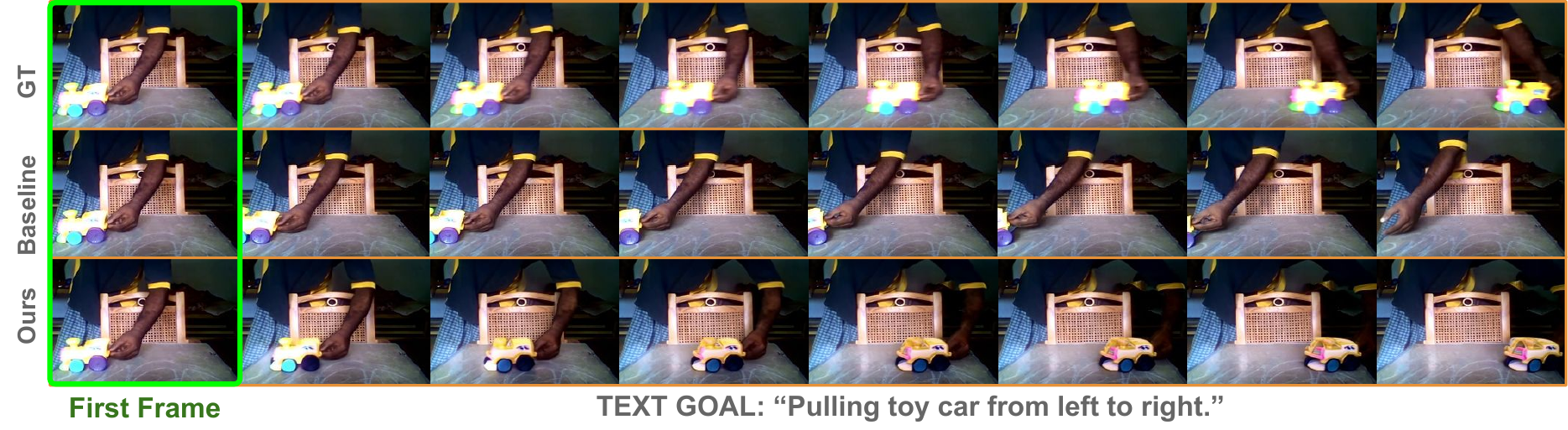}
    \includegraphics[width=\linewidth]{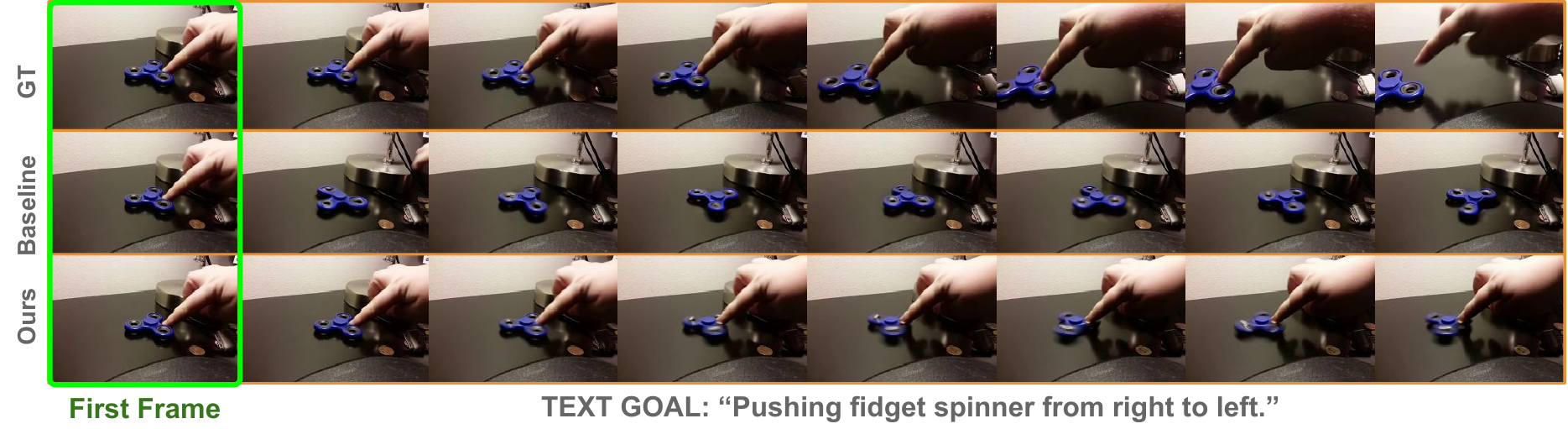}
    \includegraphics[width=\linewidth]{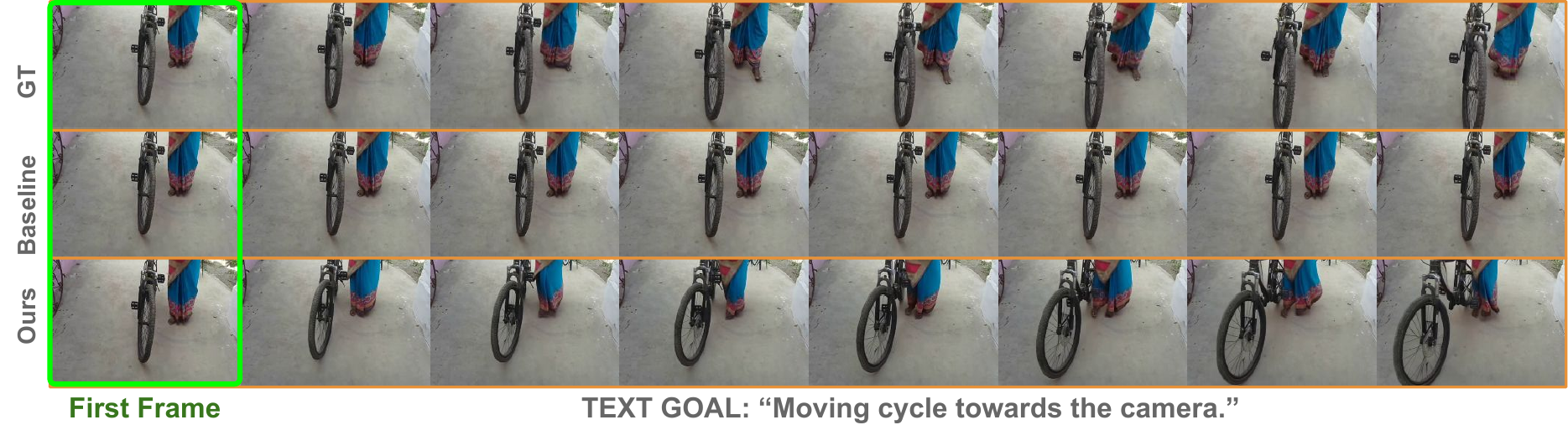}
    \caption{\textbf{Visualization of success and failure cases for Text-to-Video (T2V) generation:}
    We visualize some success and failure cases for our framework over the baseline, CogVideoX \cite{Yang2024CogVideoXTD}. 
    Examples are drawn from the SSv2 validation split. 
    We note that our method consistently improves motion adherence over the baseline. 
    However, in some cases our framework distorts the visual appearance of objects although they undergo correct movement 
    (e.g., see ``toy car'' in Row 2). 
    Checkout our \texttt{\href{https://fofpred.github.io}{FOFPred.github.io}} for more visualizations.
    }
    \label{fig:app_vis}
\end{figure*}

\section{Visualizations}
\label{app:vis}
We next visualize some videos generated with our framework for pairs of first frames and motion-focussed captions obtained from the SSv2 dataset. 
\Cref{fig:app_vis} presents this qualitative comparison between the ground truth (GT) video, our T2V baseline from CogVideoX \cite{Yang2024CogVideoXTD}, and our framework, \modelname \footnote{In our work, we explore Text-to-Video (T2V) generation using a two-stage pipeline that connects \modelname with the existing video synthesis model Go-with-the-Flow (GWTF) \cite{Burgert2025GowiththeFlowMV}. GWTF is built on top of CogVideoX~\cite{Yang2024CogVideoXTD} and extends it to additionally accept a user-provided motion prompt as input.}. The samples are selected from the SSv2 validation split to demonstrate generation capabilities conditioned on the first frame (outlined in green) and a specific text goal. As observed across the samples, \modelname consistently exhibits superior motion adherence compared to the baseline, which frequently struggles to generate significant movement or directionally accurate dynamics (e.g., the ``Moving glue stick'' and ``Moving cycle'' examples). While \modelname successfully executes the semantic requirements of the text prompts, we also visualize a partial failure case in the second row (``Pulling toy car''). In this instance, while our model correctly adheres to the motion instruction, our extended pipeline exhibits a trade-off in visual fidelity, resulting in slight distortion of the object's appearance compared to the baseline. However, the baseline in this case moves the car in the wrong direction. 

For more visualizations of our proposed \modelname framework, checkout our website \texttt{\href{https://fofpred.github.io}{FOFPred.github.io}}.


\end{document}